**С. Д. Погорілий, А. А. Крамов**

Київський національний університет імені Тараса Шевченка

проспект Академіка Глушкова, 4Г, 03022 Київ, Україна

# Метод розрахунку когерентності українського тексту

Проаналізовано основні методи автоматизованої оцінки когерентності тексту, написаного природною мовою, за допомогою підходів, основаних на машинному навчанні. Обґрунтовано доцільність використання методу графу семантичної схожості, порівняно з іншими методами. Запропоновано вдосконалення методу графу семантичної схожості за допомогою попередньої підготовки моделі, а саме здійснення навчання нейронної мережі векторного представлення речень. Проведено експериментальну перевірку роботи методу графу семантичної схожості та його модифікованих версій на множині україномовних статей наукових журналів різної тематики. Для формування навчальної та тестової вибірок, що використовуються для навчання нейронної мережі та перевірки ефективності методу відповідно, реалізовано пошукового робота для збирання даних з веб-сайтів українських наукових журналів. Виконано навчання різнотипних нейронних мереж на сформованій навчальній вибірці та здійснено програмну реалізацію побудови графу семантичної схожості для оцінки когерентності тексту. Ефективність роботи методу та його модифікацій розраховано за допомогою вирішення типових задач оцінки когерентності тексту: задач розрізнення документів та вставки. На основі отриманих результатів визначено найефективніші модифікацію та параметри методу графу семантичної схожості для оцінки когерентності україномовних текстів.

## Вступ

У зв’язку з постійним розвитком технологій машинного навчання та збільшенням обчислювальної потужності робочих станцій, спостерігається зростаючий інтерес науковців до вирішення низки задач, що неформально розглядаються як *AI-повні*. Під AI-повними задачами розуміють проблеми, для вирішення яких необхідно створити систему, що повністю відтворює розум людини [1]. Особлива увага приділяється *обробці природної мови* (англ. *Natural language processing – NLP*), про що свідчить наявність актуальних робіт та конференцій [2], пов’язаних з NLP. До задач NLP відноситься широкий спектр завдань: синтез мовлення, розпізнавання мови, машинний переклад, інформаційний пошук, спрощення тексту тощо. Задачі NLP актуальні не тільки для галузі комп'ютерної лінгвістики, але і для інших областей науки: метеорології, медицини, геофізики тощо. Активним дослідженням задач природної обробки тексту займаються науковці Стенфордського університету та корпорації Google [2–3]. Українськими науковцями теж здійснюється дослідження автоматизованої обробки текстів української мови, про що свідчить наявність корпусів, частотних словників, тезаурусів та програмних моделей обробки природної української мови [4].

Враховуючи актуальність пошукової оптимізації сайтів в мережі Internet, доцільним стає автоматизоване здійснення оцінки якості написаного тексту для покращення розуміння матеріалу читачем та підвищення позиції статті відносно конкуруючих веб-сторінок у видачі пошукової системи. До складових оцінки якості тексту відноситься поняття *когерентності*. Під когерентністю тексту розуміють цілісність тексту, що полягає в логіко-семантичній, граматичній та стилістичній співвіднесеності і взаємозалежності його складових [5]. Лінгвістами здійснюється структурний поділ поняття когерентності на три види: локальну, глобальну та тематичну когерентності. *Локальна когерентність* визначається міжфразовими синтаксичними зв'язками; *глобальна* забезпечує єдність тексту як смислового цілого, його внутрішню цілісність. *Тематична когерентність* проявляється за допомогою ключових слів, які тематично поєднують фрагменти тексту. В цілому, когерентність тексту дозволяє оцінити тематичну узгодженість складових частин тексту, його смислову цілісність, і, відповідно, зрозумілість матеріалу тексту читачу. Питання оцінки когерентності тексту активно розглядається науковцями: запропоновано низку методів для знаходження міри цілісності тексту за допомогою машинних методів навчання. Однак основний вектор актуальних досліджень напрямлений на оцінку когерентності текстів англійської мови; для української мови, що має свої морфологічні, синтаксичні та семантичні особливості, необхідно здійснювати індивідуальне дослідження. Тому актуальною є задача оцінки когерентності текстів української мови за допомогою існуючих методів, а також їх подальше вдосконалення відповідно до особливостей мови.

Метою цієї роботи є порівняльний аналіз існуючих методів оцінки когерентності тексту природної мови; вдосконалення певного методу та здійснення експериментальної перевірки ефективності роботи різних модифікацій обраного методу для корпусу української мови.

**Огляд існуючих методів оцінки когерентності тексту**

Вхідними даними кожного методу є текст природної мови. Результатом роботи є числове значення – міра когерентності тексту. Розглянемо основні методи оцінки цілісності тексту, після чого зупинимося на більш детальному огляді певного методу.

У 2008 році була запропонована модель оцінки когерентності тексту, яка отримала назву **Entity Grid** [6]. Головна ідея моделі полягає в припущенні, що розподіл ключових сутностей тексту (іменні групи, які присутні в реченнях) підпорядковується певній закономірності. Параметром оцінки когерентності вважається частота зміни ролі ключових сутностей в тексті (підмет, прямий додаток, інше), тобто аналізується частота зміни наголосів в тексті, які привертають увагу читача. У випадку різких/рівномірних переходів від однієї ключової сутності до іншої оцінка когерентності відповідно зменшується/збільшується. Для навчання моделі автори використовували метод опорних векторів (support vector machine – SVM). Таким чином, результат методу Entity Grid є дискретним значенням: 1, якщо текст когерентний; 0 в іншому випадку.

В 2013 році був запропонований новий метод оцінки когерентності тексту, який отримав назву **Entity Graph** [7]. У методі Entity Graph здійснюється побудова графу, що інтерпретує текст, для оцінки взаємозв’язку між сусідніми та віддаленими реченнями (на відміну від Entity Grid, де переважно розглядався локальний зв’язок речень). Текст представлений за допомогою орієнтованого двочасткового графу (біграфу). Перша підмножина вершин графу відповідає набору речень тексту. У другій підмножині графу кожній вершині ставиться у відповідність сутність. Зважене ребро між вершинами речення і сутності встановлюється у разі наявності сутності в реченні. Вага ребра встановлюється залежно від ролі сутності в реченні (підмет, прямий додаток, без ролі). Отриманий орієнтований двочастковий граф далі перетворюється в орієнтований проекційний граф, в якому всі вершини позначають речення тексту, а ребра між ними встановлюються за умови наявності щонайменше однієї спільної сутності. Напрямок ребер відповідає послідовності розміщення речень у тексті. Ваги ребер розраховуються відповідно до трьох різних проекційних методів. Для оцінки когерентності тексту обраховується середнє арифметичне значення напівстепені виходу вершин графу.

Існують методи, що використовують різнотипні **нейронні мережі** для оцінки когерентності тексту. Відповідні методи реалізовані на основі рекурсивної і рекурентної [8] та згорткової нейронних мереж [9]. Основна ідея полягає у початковому розбитті тексту на речення та подальшому перетворенні «речення» – «вектор»; таку функцію виконують початкові шари нейронних мереж. Варто зазначити, що для такого перетворення використовуються заздалегідь навчені моделі векторного представлення слів: Word2Vec, GloVe та їх модифікації. Після здійснення перетворення «речення» – «вектор» наступними шарами нейронних мереж виконується пошук закономірностей між векторами речень згідно зі структурою мережі та значеннями ваг ребер. Вихідним результатом є числове значення; чим більше числове значення, тим вища оцінка когерентності. Недоліком є відсутність нормалізації вихідного значення та непрозорість моделі: неможливо відслідкувати причину встановлення низької/високої оцінки когерентності вхідного тексту.

Останнім розглянемо метод **графу семантичної схожості** (ГСС) [10]. Як і Entity Graph, ГСС відноситься до графічних методів оцінки когерентності тексту. Спочатку текст трансформується до орієнтованого графу $G(V, E)$, де $V$ – множина вершин, $E$ – множина ребер графу. Вершина $v_i \in V$ відповідає $i$-тому реченню $s_i$ в тексті, а зважене напрямлене ребро $e_{ij} \in E$ описує міру семантичної схожості між реченнями $s_i$ та $s_j$. Представлення речення здійснюється у векторному вигляді. Для цього кожне слово речення трансформується у вектор за допомогою попередньо навченої моделі Word2Vec чи GloVe. Таким чином, речення складається з множини слів $\{w_1, w_2, \dots, w_M\}$ ($M$ – кількість слів речення), де кожному слову відповідає вектор $\{\mathbf{w}_1, \mathbf{w}_2, \dots, \mathbf{w}_M\}$. Речення $s$ може бути представлене як вектор $\mathbf{s}$ за допомогою усереднення векторів його слів:

$$\mathbf{s} = \frac{1}{M}\sum_{k=1}^{M} \mathbf{w}_k \quad (1)$$

Для формування графу пропонується використовувати три різних підходи: PAV, SSV, MSV. У підході *PAV* (preceding adjacent vertex – попередня сусідня вершина) здійснюється формування ребер від вершини кожного речення до вершини сусіднього попереднього речення, якщо міра схожості між цими реченнями більша 0; в іншому випадку здійснюється спроба встановити ребро між поточною та іншою найближчою попередньою вершинами. Міра схожості речень розраховується за наступною формулою:

$$\mathrm{sim}(s_i, s_j) = \alpha \mathrm{uot}(s_i, s_j) + (1 - \alpha)\cos(\mathbf{s}_i, \mathbf{s}_j), \quad (2)$$

де $\mathrm{uot}$ – відношення кількості спільних сутностей речень $s_i$ та $s_j$ до загальної кількості сутностей цих речень; $\cos(\mathbf{s}_i, \mathbf{s}_j)$ – косинусна відстань між векторами речень; $\alpha$ – регулятивний параметр, $\alpha \in [0,1]$.

На відміну від PAV, у підході *SSV* (single similar vertex – єдина схожа вершина) залежність між реченнями може мати двонаправлений характер, тому ребра можуть бути направлені не тільки на вершини попередніх речень. Однак зберігається обмеження значення напівстепені виходу кожної вершини, не більшого за 1.

Для встановлення вихідного ребра поточної вершини здійснюється пошук речення, яке є найбільш схожим до поточного в семантичному розумінні. Міра схожості, відповідна вазі ребра, розраховується як відношення косинусної відстані між векторами речень до відстані між реченнями в тексті:

$$\text{weight}(e_{ij}) = \frac{\cos(\mathbf{s}_i, \mathbf{s}_j)}{|i-j|} \qquad (3)$$

У попередніх підходах PAV і SSV кожна вершина була або інцидентна одному ребру, направленому до іншої вершини, або ізольованою. У підході *MSV* (multiple similar vertex – багато схожих вершин) це обмеження ігнорується: для кожного речення здійснюється пошук всіх речень, міра схожості (див. формулу (3)) яких з поточним реченням більша за порогове значення $\theta$. Ваги сформованих ребер рівні мірі схожості суміжних вершин. Приклади всіх підходів до формування ГСС зображено на рис. 1.

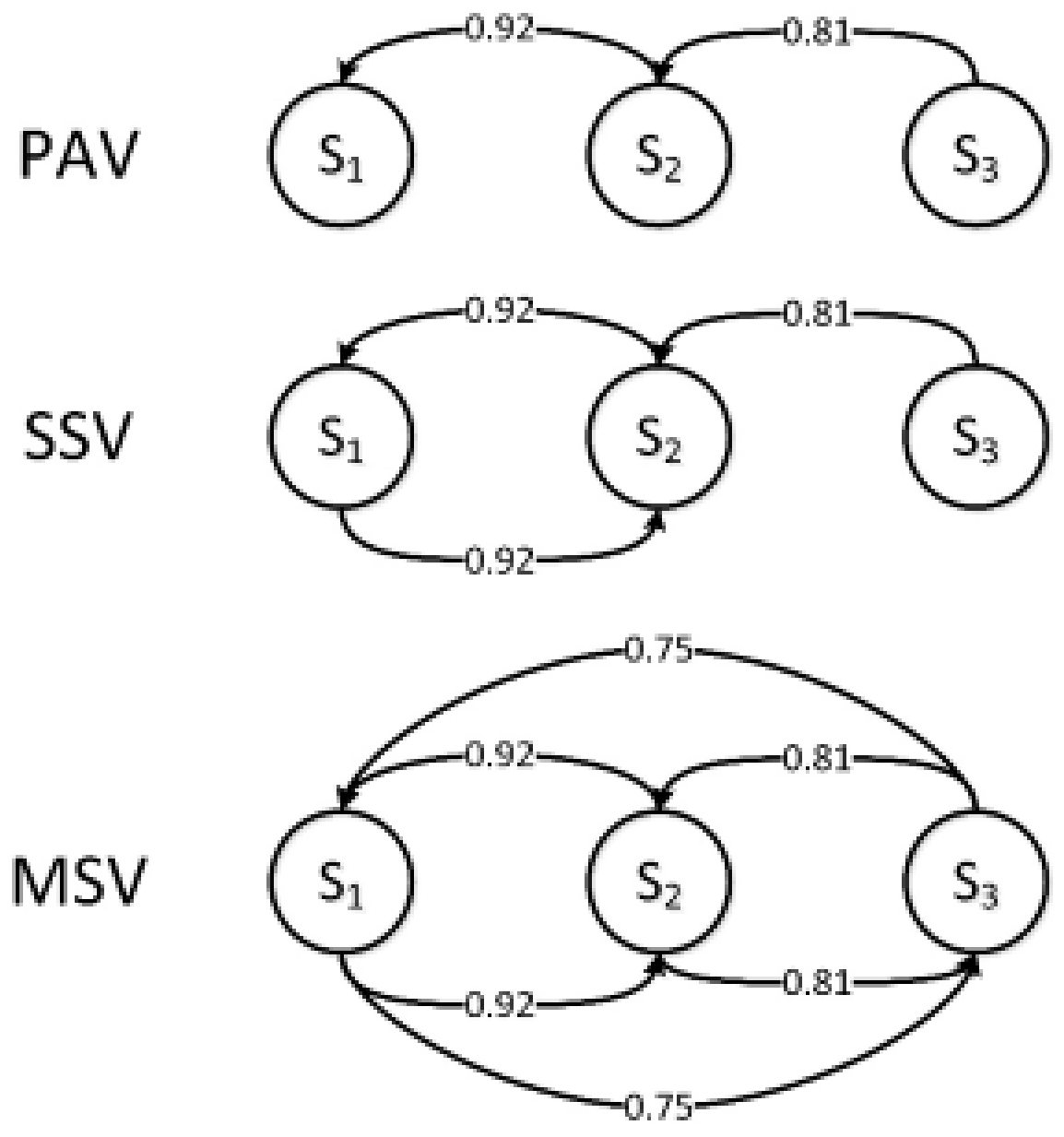

Рис. 1 – Приклади підходів до побудови ГСС

Оцінка когерентності тексту $t_c$ здійснюється за рахунок аналізу сформованого графу і обраховується наступним чином:

$$t_c = \frac{1}{N}\sum_{i=1}^{N}\frac{1}{L_i}\sum_{k=1}^{L_i}\text{weight}(e_{ik}), \qquad (4)$$

де $N$ – кількість речень в тексті; $L_i$– кількість ребер, які виходять з вершини $v_i$. Значення $t_c$ нормалізоване (приймає значення з діапазону $[0,1]$) і прямо пропорційне мірі когерентності тексту.

Графічні методи оцінки когерентності тексту (ГСС, Entity Graph) дозволяють водночас відстежити процес формування вихідного результату методу, а також проаналізувати тематичну схожість певних речень для подальшого покращення цілісності тексту. Моделі інших методів розглядаються як «чорний ящик», що унеможливлює аналіз причини формування вихідної оцінки когерентності тексту та виконання її підвищення. Особливістю методу ГСС, порівняно з Entity Graph, є аналіз власне семантичної складової тексту: Entity

Graph використовує синтаксичний аналіз іменованих сутностей для відстеження зміни їх ролей. Також до переваг ГСС варто віднести неперервність та нормалізацію вихідної оцінки когерентності тексту. Таким чином, проаналізувавши основні методи оцінки когерентності тексту, було вирішено використати метод ГСС для здійснення експериментальної перевірки ефективності роботи методів на корпусі української мови.

**Модифікація методу ГСС**

Принцип роботи методу ГСС полягає в побудові орієнтованого зваженого графу, вершини якого представляють речення тексту, а ваги ребер – семантичну схожість речень. Когерентність оцінюється як середнє арифметичне значення ваг ребер графу. Нетривіальною виглядає задача знаходження міри семантичної схожості речень. Міра схожості речень розраховується відповідно до обраного підходу (PAV, SSV, MSV) за допомогою попереднього розрахунку векторного представлення речень. Вектор кожного речення обчислюється як середнє арифметичне векторів слів, що входять до речення (див. формулу (1)). Для векторного представлення слів попередньо виконується навчання нейронної мережі Word2Vec або GloVe.

Такий підхід до векторного представлення речень використовувався в багатьох роботах, пов'язаних з машинною обробкою тексту, та виявився доволі ефективним. Однак описаний підхід має декілька недоліків, а саме:

- неможливо здійснювати векторне представлення слова, незнайомого навченій моделі, без додаткового етапу навчання;
- не вирішується проблема омонімії, адже кожне слово має єдине векторне представлення незалежно від контексту;
- не враховується порядок слів у тексті.

Відсутність слова у словнику навченої моделі обумовлює або ігнорування слова в тексті, або пошук найбільш схожого слова. Зрозуміло, що така апроксимація погіршує семантичне представлення речення в цілому. Ігнорування порядку слів у тексті може призвести до некоректного смислового навантаження речень (варто зазначити, що для різних мов фактор впливу порядку слів у реченні відрізняється). Крім того, векторне представлення слова не вирішує проблему омонімії, адже словам, які однаково звучать та пишуться, але мають різне значення, відповідатиме спільний вектор; таким чином не враховується контекст вживання слова в реченні.

Вирішити описані вище проблеми пропонується за допомогою навчання нейронної мережі векторного представлення документів – Doc2Vec [11]. Навчена модель Doc2Vec дозволяє здійснювати векторне представлення групи слів, речень, абзаців тощо (залежно від конкретної задачі). У випадку знаходження семантичної схожості речень доцільно здійснювати навчання моделі на колекціях речень. Процес навчання нейронної мережі може відбуватися за допомогою підходу «навчання без вчителя», що дозволяє використовувати нерозмічені дані та виконувати векторне представлення речень, незнайомих навченій моделі.

В роботі [11] описано дві моделі Doc2Vec: Distributed Bag of Words (DBOW) і Distributed Memory (DM). Відмінність моделей полягає в наступному: модель DM, на відміну від DBOW, при навчанні враховує порядок слів речення. Це зумовлює більше використання пам'яті моделлю DM та повільнішу роботу, порівняно з методом DBOW. Можливим є здійснення об'єднання моделей DBOW і DM (далі DBOW+DM), хоча таку комбінацію моделей методу Doc2Vec варто розглядати як експериментальну.

**Експериментальна перевірка роботи методу ГСС**

Експериментальну перевірку роботи методу ГСС варто розділити на наступні етапи:

- генерація навчального корпусу та тестової вибірки;
- навчання моделей нейронних мереж;
- програмна реалізація різнотипних моделей ГСС;
- тестування ефективності моделей у вирішенні типових задач оцінки когерентності тексту.

**Генерація навчального корпусу та тестової вибірки**

Першим етапом експериментальної перевірки роботи методу ГСС стало формування навчальної вибірки для нейронних мереж Word2Vec, DBOW і DM. Внаслідок пошуку існуючих корпусів української мови було знайдено готові рішення, створені відкритою спільнотою фахівців lang-uk [4]. Представники спільноти lang-uk надали відкритий доступ до попередньо оброблених корпусів української мови, сформованих з новин, творів художньої літератури, статей ресурсу «Вікіпедія». Однак тексти, розміщені в корпусах, представлені у вигляді перемішаних речень згідно з вимогами правовласників текстів. Для тестування методу ГСС необхідні оригінальні версії текстів, тому було прийняте рішення сформувати навчальну вибірку власноруч. Як вхідні дані було вирішено обрати публікації українських наукових журналів. На сайтах наукових журналів повний текст публікацій доступний лише в PDF-форматі, що значно ускладнює процес формування навчальної вибірки, адже необхідні додаткові операції екстракції власне тексту статті в належному форматі. Тому для формування навчального корпусу було вирішено використати анотації статей, які можливо отримати з HTML-розмітки. Враховуючи, що анотації містять основні тематичні терміни, які стосуються відповідних статей, такий підхід повинен дозволити водночас навчити моделі працювати з реченнями, які стосуються певної тематики, а також зменшити об'єм навчальної вибірки. Для перевірки цього припущення, як тестовий набір обрано повні версії статей, екстраговані з PDF-файлів.

Генерацію навчального корпусу та тестового набору даних виконано окремим консольним застосуванням, написаним мовою програмування Python 3.6. Збір HTML-сторінок та PDF-файлів виконано за допомогою реалізації пошукового робота, який працює як HTTP-клієнт [12–13]. Було здійснено програмний обхід 266 веб-сайтів українських наукових журналів, які відносяться до категорій «Природничі та точні науки» і «Соціогуманітарні науки». Значна кількість статей, розміщених на проаналізованих веб-ресурсах, написана російською або англійською мовами, тому додатково було виконано автоматизоване детектування української мови програмним пакетом *langdetect,* розробленим корпорацією Google.

Наступним кроком генерації навчального корпусу було виконання попередньої обробки текстів, а саме:

- розбиття тексту на речення;
- вилучення стоп-слів;
- здійснення лематизації слів.

Вилучення стоп-слів з речень виконане за допомогою ітераційної перевірки наявності слова в списку стоп-слів української мови. Лематизація – це процес перетворення слова у словниковий вид або лему. Іменники та прикметники лематизація трансформує в форму однини, дієслово – в інфінітив (відповідь на питання «що робити?»). На жаль, для мови програмування Python відсутні пакети лематизації української мови. Єдиним рішенням, знайденим для української мови, стали утиліти, створені попередньо зазначеною відкритою спільнотою фахівців lang-uk. Утиліти написані мовою програмування Groovy, тому було створено «обгортку» мовою програмування Python для виконання утиліт у паралельному потоці програми. За допомогою утиліт було здійснено водночас розбиття тексту на речення та лематизація слів. Для створення навчальної вибірки було використано 74 180 анотацій; загальна кількість речень для навчання моделі становить 355 537.

Для формування тестового набору текстів необхідно було виконати екстракцію матеріалу статті з PDF-файлу. Існуючих рішень для українських наукових публікацій не було знайдено у відкритому доступі, тому було вирішено використати інше рішення – клієнт-серверне застосування Science Parse [14]. В результаті було сформовано близько 1000 тестових текстів.

**Навчання моделей нейронних мереж**

Навчання нейронних моделей Word2Vec, DBOW і DM на сформованій навчальній вибірці було виконано за допомогою окремого застосування, написаного мовою програмування Python 3.6. Для виконання навчання було використано готові класи *Word2Vec* і *Doc2Vec* пакету *genism* [15]. Для прискорення процесу навчання було використано розширення Cython. Навчання здійснювалося з наступними параметрами:

- розмірність вектору – 300;
- кількість епох – 50;
- розмір «вікна» – 10;
- порогове значення частотного словника – 1.

Запуск застосування було здійснено на робочій станції з наступними характеристиками: Intel Core i7-7700 (3.6 – 4.2 ГГц) / RAM 32 ГБ / SSD. Результати навчання нейронних мереж – моделі Word2Vec, DBOW, DM – збережено на файловій системі з можливістю їх завантаження в пам’ять програми та подальшого використання.

**Програмна реалізація різнотипних моделей ГСС**

Програмна реалізація ГСС здійснена мовою програмування Python 3.6. Застосування містить 3 базових класи, кожен з яких реалізує відповідний підхід до побудови ГСС: PAV, SSV, MSV. Базові класи наслідують батьківський клас, в якому описані загальні функції для дочірніх класів. Для кожного класу передбачено можливість змінити модель (для здійснення векторного представлення речень) і функцію обрахунку міри семантичної схожості речень. Для графів PAV і MSV спроектовано можливість змінювати параметри налаштування: регулятивний параметр $\alpha$ для PAV і порогове значення $\theta$ для MSV. Результат роботи вказаних підходів залежить від обраних значень параметрів, що потребує подальшої експериментальної перевірки.

**Тестування**

**Задача розрізнення документів**

Задача розрізнення документів (англ. document discrimination task – DDT) полягає в наступному: для документу з перевірочної множини обраховується його когерентність; потім речення тексту випадковим чином перемішуються і обраховується когерентність зміненого тексту. Якщо значення когерентності оригіналу більше значення модифікованого тексту, вважається, що текст оброблений вірно. Оцінка задачі розрізнення документів $S_{DDT}$ розраховується наступним чином:

$$S_{DDT} = \frac{N_{correct}}{N_{total}}, \quad (5)$$

де $N_{correct}$ – кількість вірно оброблених документів; $N_{total}$ – загальна кількість документів.

Для ГСС PAV задача розрізнення документів виконувалася для різних значень регулятивного параметру $\alpha$ з кроком $0.1$. Аналогічно здійснювалося тестування для порогового значення $\theta$ ГСС MSV. В табл. 1 наведені максимальні результати вирішення задачі розрізнення документів методом ГСС з відповідними підходами та їх параметрами для тестової множини документів. На відміну від підходу SSV, результати роботи PAV і MSV залежать від параметрів налаштування. Проаналізуємо більш детально залежність ефективності зазначених підходів від зміни параметру. На рис. 2 зображено динаміку зміни оцінки задачі розрізнення документів відносно регулятивного параметру $\alpha$ для підходу PAV. На графіку спостерігається поступове підвищення точності роботи підходу PAV зі збільшенням регулятивного параметру; така залежність вказує на необхідність врахування різноманітних елементів міжфразової єдності (синонімів, антонімів, кореферентів) при здійсненні оцінки когерентності текстів української мови. Щодо порівняльної характеристики моделей, різні варіанти Doc2Vec виявилися ефективнішими, порівняно з моделлю Word2Vec. Найвищі результати отримані за допомогою конкатенації моделей DBOW+DM (проігноровано значення моделі Word2Vec при $\alpha = 0$ у зв'язку зі значним відхиленням від множини інших отриманих значень моделі), однак необхідно зазначити наявність непропорційних змін вихідних результатів відносно регулятивного параметру в моделях DBOW і DBOW+DM. На рис. 3 зображено графік зміни результату задачі розрізнення документів відносно порогового значення $\theta$ для підходу MSV. На графіку простежується зменшення ефективності роботи кожної моделі при збільшенні порогового значення. Така залежність вказує на необхідність врахування зв'язку між всіма реченнями тексту, адже максимальний результат спостерігається при пороговому значенні $\theta = 0$. Найкращий результат вирішення задачі розрізнення документів отриманий за допомогою моделі DM. Незважаючи на високу ефективність комбінованої моделі DBOW+DM, порівняно з іншими моделями, для підходів PAV і SSV, її результати роботи для підходу MSV виявилися найнижчими.

| Підхід | Параметр | Задача розрізнення документів | | Задача вставки | |
|---|---|---|---|---|---|
| | | Модель | Значення | Модель | Значення |
| PAV | $\alpha = 0.8$ | DBOW | 0.661 | DM | 0.532 |
| SSV | – | DBOW+DM | 0.628 | DBOW+DM | 0.227 |
| MSV | $\theta = 0$ | DM | 0.808 | DM | 0.76 |

Таблиця 1 – Максимальні результати вирішення задач розрізнення документів і вставки методом ГСС

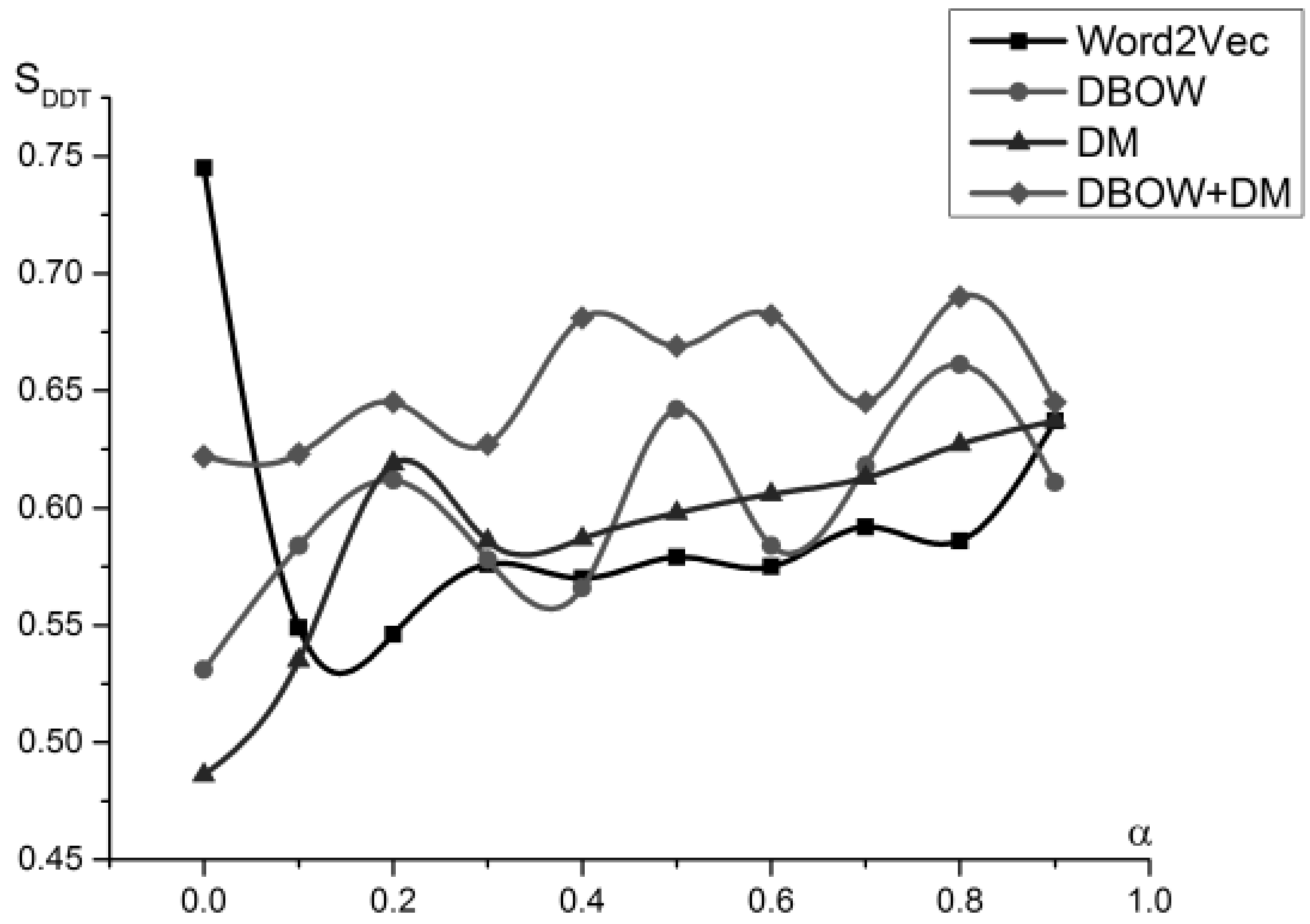

Рис. 2 – Динаміка зміни оцінки задачі розрізнення документів відносно регулятивного параметру $\alpha$ для підходу PAV

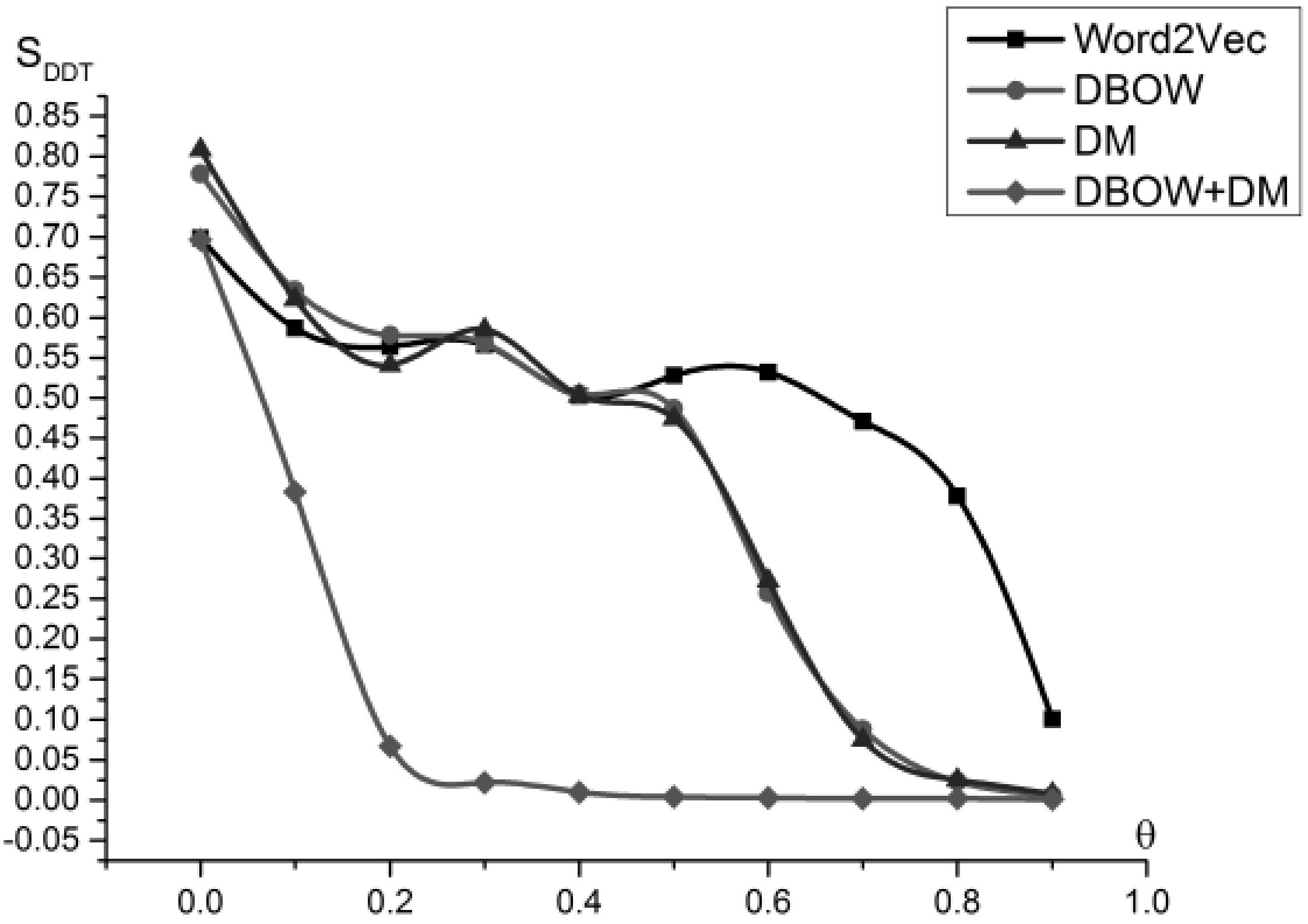

Рис. 3 – Динаміка зміни оцінки задачі розрізнення документів відносно порогового значення $\theta$ для підходу MSV

**Задача вставки**

Алгоритм дій для виконання задачі вставки (англ. insertion task – IT) наступний: з тексту випадковим чином вилучається одне речення; потім видалене речення вставляється почергово у всі можливі позиції тексту. Для кожного варіанту вставки обраховується когерентність тексту. Обирається варіант вставки з найбільшим значенням когерентності. Якщо обраний варіант вставки співпадає з оригінальним текстом, тоді задача вставки вважається виконаною вірно для цього тексту. Оцінкою задачі вставки є значення $S_{IT}$, що розраховується за наступною формулою:

$$S_{IT} = \frac{N_{correct}}{N_{total}}, \quad (6)$$

де $N_{correct}$ – кількість вірно оброблених документів; $N_{total}$ – загальна кількість документів. В табл. 1 наведені максимальні результати вирішення задачі вставки методом ГСС для тестової множини документів. Розглянемо детальніше залежності підходів PAV і MSV від параметрів налаштування. На рис. 4 зображено графік залежності оцінки задачі вставки від значення регулятивного параметру $\alpha$ для підходу PAV. Порівняно з моделлю Word2Vec, всі інші моделі Doc2Vec виявилися ефективнішими. Найкращий результат було отримано для моделі DM. Аналогічно до задачі розрізнення документів, на графіку спостерігається покращення ефективності роботи підходу PAV зі збільшенням регулятивного параметру $\alpha$.

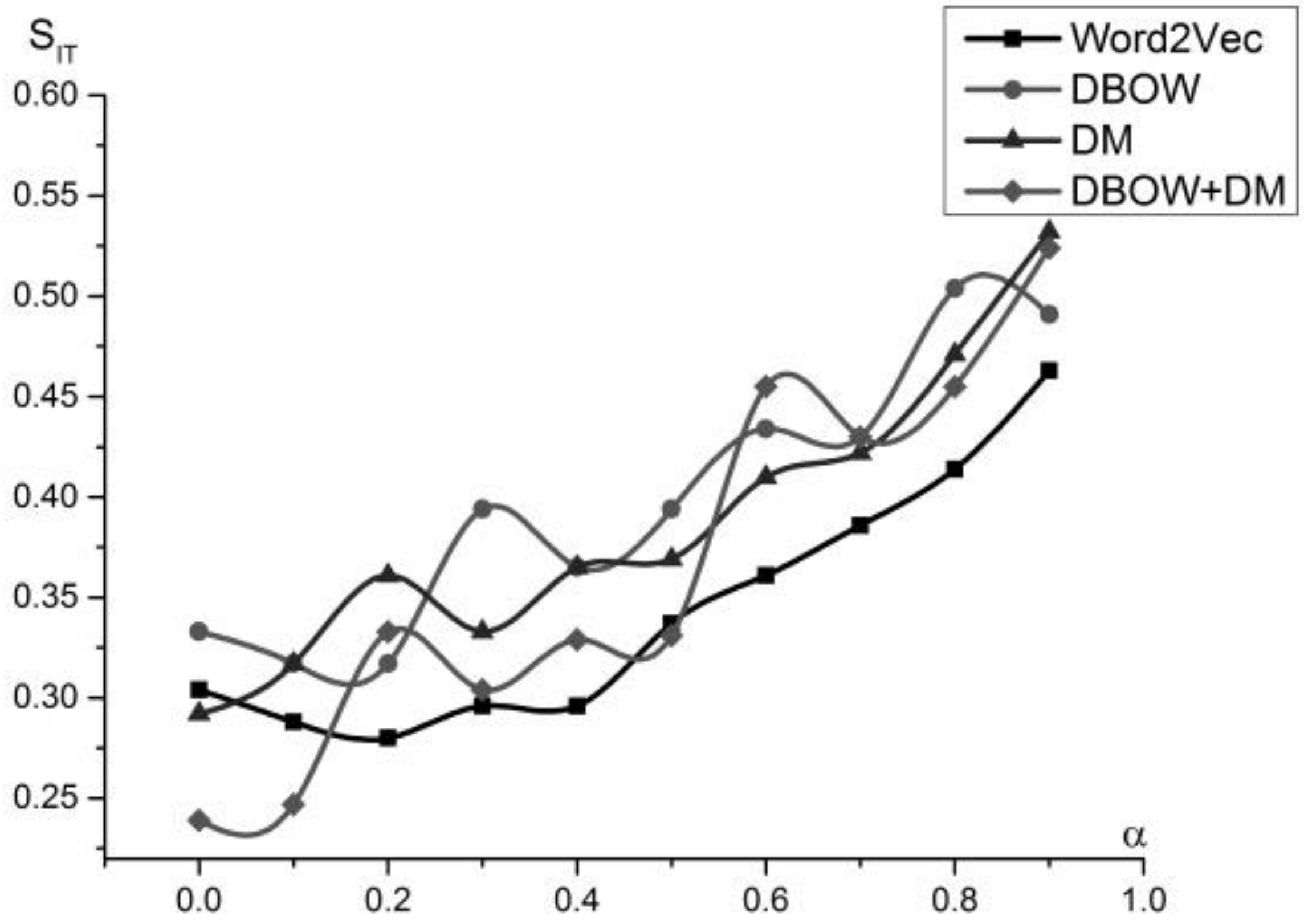

Рис. 4 – Динаміка зміни оцінки задачі вставки відносно регулятивного параметру α для підходу PAV

На рис. 5 зображено графік зміни оцінки задачі вставки відносно порогового значення $\theta$ для підходу MSV. Аналогічно до задачі розрізнення документів найефективнішою виявилася модель DM, а комбінація моделей DBOW+DM показала найнижчі результати. Результати, отримані для задачі вставки за допомогою підходу MSV, підтверджують необхідність розгляду вхідного тексту як насиченого графу: кожна вершина, що представляє речення, суміжна з будь-якою іншою вершиною графу.

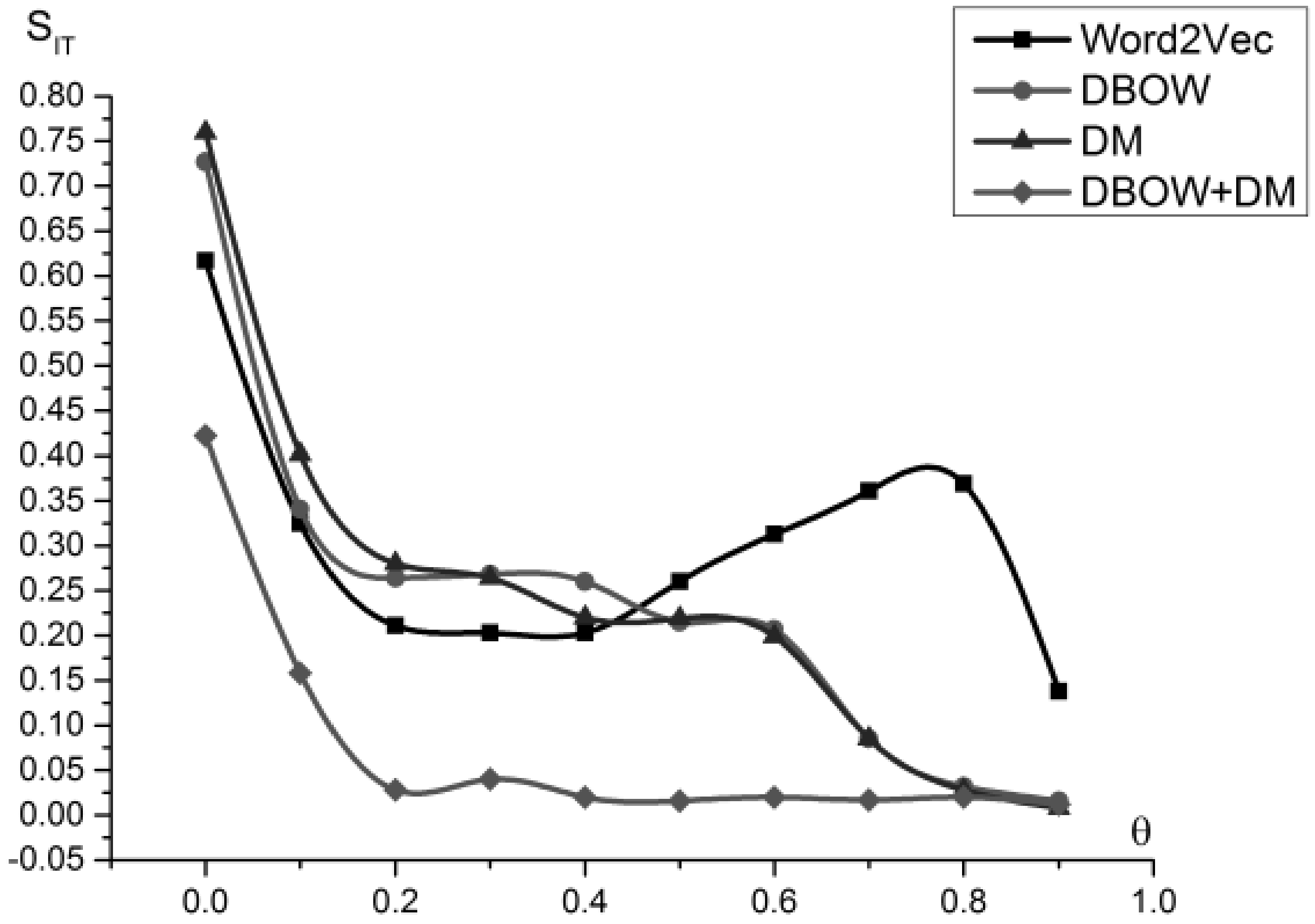

Рис. 5 – Динаміка зміни оцінки задачі вставки відносно порогового значення θ для підходу MSV

**Висновки**

На основі аналізу отриманих результатів, можна зробити такі висновки:

- найефективнішою комбінацією моделі, підходу до побудови графу і його параметрів є граф семантичної схожості MSV з пороговим значенням $\theta = 0$, що використовує модель DM;
- граф семантичної схожості SSV показав найгірші результати для типових задач оцінки ефективності роботи методу розрахунку когерентності тексту, що вказує на наявність багаторазового зв'язку між реченнями тексту, написаного природною українською мовою;

- високі показники роботи графу PAV для значення параметру $\alpha > 0.5$ вказують на доцільність врахування спільних термів речень;
- динаміка спаду результатів зі збільшенням порогового значення $\theta$ для графу MSV вказує на необхідність врахування зв'язку між всіма реченнями тексту, незалежно від відстані між ними; відсутність поступового зменшення результатів для моделі Word2Vec можна пояснити фактором випадкового вибору вектору для слів, відсутніх у словнику навченої моделі;
- комбінація моделей DBOW+DM виявилася ефективною лише для графів PAV і SSV, що враховують одноразовий зв'язок між реченнями; такий результат можна спробувати трактувати як перенавчання комбінованої моделі, адже модель нездатна враховувати зв'язки між реченнями, розташованими не поруч одне з одним;
- наявність багаторазового зв'язку та спільних термів між реченнями дозволяють зробити висновок про доцільність врахування *елементів міжфразової єдності* (синоніми, антоніми, гіпоніми, кореферентні зв'язки) для покращення результатів роботи методу ГСС, що і пропонується виконати в наступних дослідженнях.

**Перелік посилань**

1. Raymond E. S. The new hacker's dictionary. Mit Press, 1996. 568 p.
2. Publications - The Stanford Natural Language Processing Group. URL: https://nlp.stanford.edu/pubs (дата звернення: 08.12.2018).
3. Publications – Google AI. URL: https://ai.google/research/pubs (дата звернення: 08.12.2018).
4. Homepage: lang-uk. URL: http://lang.org.ua (дата звернення: 02.12.2018).
5. Лєднік О. С. Когезія та когерентність як категорії зв'язного тексту. *Науковий часопис Національного педагогічного університету імені М. П. Драгоманова. Серія 10: Проблеми граматики і лексикології української мови*. 2010. Вип. 6. С. 119–123.
6. Barzilay R., Lapata M. Modeling local coherence: An entity-based approach. *Computational Linguistics*. 2008. Vol. 34, No 1. P. 1–34.
7. Guinaudeau C., Strube M. Graph-based local coherence modeling. *Proceedings of the 51st Annual Meeting of the Association for Computational Linguistics*. 2013. Vol. 1. P. 93–103.
8. Li J., Hovy E. A model of coherence based on distributed sentence representation. *Proceedings of the 2014 Conference on Empirical Methods in Natural Language Processing (EMNLP)*. 2014. P. 2039–2048.

9. Cui B., Li Y., Zhang Y., Zhang Z. Text Coherence Analysis Based on Deep Neural Network. *Proceedings of the 2017 ACM on Conference on Information and Knowledge Management*. 2017. P. 2027–2030.
10. Putra J. W. G., Tokunaga T. Evaluating text coherence based on semantic similarity graph. *Proceedings of TextGraphs-11: the Workshop on Graph-based Methods for Natural Language Processing*. 2017. P. 76–85.
11. Le Q., Mikolov T. Distributed representations of sentences and documents. *International Conference on Machine Learning*. 2014. P. 1188–1196.
12. Погорілий С. Д., Крамов А. А. Автоматизована екстракція структурованої інформації з множини веб-сторінок. *Проблеми програмування.* 2018. № 2–3. С. 149–158.
13. Pogorilyy S., Kramov A. Automated extraction of structured information from a variety of web pages. *Proceedings of the 11th International Conference of Programming UkrPROG 2018*. Kyiv, 2018. P. 149–158.
14. Science Parse Server. URL: https://github.com/allenai/science-parse/blob/master/server/README.md (дата звернення: 08.12.2018).
15. gensim: Topic modelling for humans. URL: https://radimrehurek.com/gensim (дата звернення: 08.12.2018).

**Відомості про авторів**

*Погорілий Сергій Дем'янович*

Науковий ступінь: доктор технічних наук.

Вчене звання: професор.

Посада: завідувач кафедри комп'ютерної інженерії факультету радіофізики, електроніки та комп'ютерних систем Київського національного університету імені Тараса Шевченка.

Установа: Київський національний університет імені Тараса Шевченка.

Домашня адреса: 01135, Київ, вул. В'ячеслава Чорновола, 10, кв.109.

Службова адреса: 03022, Київ, проспект Академіка Глушкова, 4Г.

Телефон: +38 (066) 434-27-86

E-mail: sdp77@i.ua

*Крамов Артем Андрійович*

Науковий ступінь: –.

Вчене звання: –.

Посада: аспірант кафедри комп'ютерної інженерії факультету радіофізики, електроніки та комп'ютерних систем Київського національного університету імені Тараса Шевченка.

Установа: Київський національний університет імені Тараса Шевченка.

Домашня адреса: 38800, смт. Чутове, Полтавська область, вул. Полтавський Шлях, 87.

Службова адреса: 03022, Київ, проспект Академіка Глушкова, 4Г.

Телефон: +38 (050) 149-31-32

E-mail: artemkramov@gmail.com